\documentclass{article}
\usepackage{spconf,amsmath,graphicx,amsfonts}
\usepackage{kotex}
\usepackage{xcolor}

\newcommand{\eg}{\textit{e}.\textit{g}.}

\title{Continuous Face Aging Generative Adversarial Networks}
\name{Seogkyu Jeon$^1$, Pilhyeon Lee$^1$, Kibeom Hong$^1$, and Hyeran Byun$^{1,2*}$
\thanks{* Corresponding Author\newline \hspace*{1em} This work was partly supported by the National Research Foundation of Korea (NRF) grant funded by the Korea government (MSIT) (No.2019R1A2C2003760) and Institute for Information \& communications Technology Planning \& Evaluation (IITP) grant funded by the Korea government (MSIT) (No.2020-0-01361, Artificial Intelligence Graduate School Program (YONSEI UNIVERSITY)).}}
\address{$^1$Department of Computer Science, Yonsei University \\
$^2$Graduate School of Artificial Intelligence, Yonsei University \\
\small \tt \{jone9312, lph1114, cha2068, hrbyun\}@yonsei.ac.kr}
\begin{document}
%\ninept
%
\maketitle
\begin{abstract}
Face aging is the task aiming to translate the faces in input images to designated ages.
To simplify the problem, previous methods have limited themselves only able to produce discrete age groups, each of which consists of ten years.
% This limits them only able to produce discrete results for the pre-defined target age groups.
Consequently, the exact ages of the translated results are unknown and it is unable to obtain the faces of different ages within groups.
To this end, we propose the continuous face aging generative adversarial networks (CFA-GAN).
Specifically, to make the continuous aging feasible, we propose to decompose image features into two orthogonal features: the identity and the age basis features.
% They are in turn trained to contain respective information and allow for handling the continuous age attributes.
Moreover, we introduce the novel loss function for identity preservation which maximizes the cosine similarity between the original and the generated identity basis features.
With the qualitative and quantitative evaluations on MORPH, we demonstrate the realistic and continuous aging ability of our model, validating its superiority against existing models.
To the best of our knowledge, this work is the first attempt to handle continuous target ages.

\end{abstract}
\section{Introduction}
\label{sec:intro}

Someone would imagine how the appearance of people changes as time goes forward or backward.
% another person's future or past.
To make this possible in reality, the face aging problem has been actively studied, which aims at translating a facial image into an older (or younger) facial image while preserving the personal identity.
Face aging has drawn much attention due to its broad use in photo-editing~\cite{fu2010age} or finding missing children~\cite{park2010age}.

\begin{figure*}[t]
 \centering
\includegraphics[width=\textwidth]{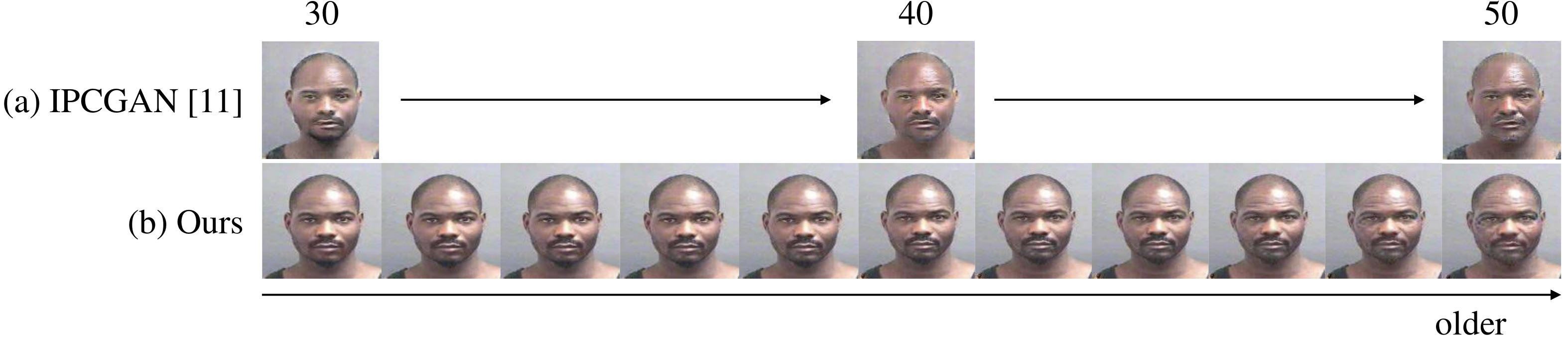}
%  \vspace{2.0cm}
% \end{minipage}
\caption{Translated results from the existing method and our CFA-GAN. Target age groups are provided above the images. Previous face aging method (IPCGAN~\cite{IPCGAN}) groups age attributes to simplify the problem. This makes it not able to generate different images of continuous target ages within each age group (a). In contrast, our CFA-GAN generates smooth and continuous aging results along with ages with interval of 2 (b). We note that the original input image is at age 34.}
\label{fig:intro_figure}
\end{figure*}

Recent studies on face aging~\cite{wang2016recurrent, antipov2017face, palsson2018generative, liu2019attribute, he2019s2gan} utilize the image-to-image translation techniques~\cite{pix2pix, cyclegan},
% The main advantage of the image-to-image translation is
thanks to its ability of translating source images to the target domain while preserving the context.
% To translate between different domains, model need to understand features of intra-domain and inter-domain relations. 
% In face aging, since its goal lies not only in translating aging factors but also in preserving personal identities, image-to-image translation can handle the objectives as domains and contexts respectively.
However, directly mapping face aging to image-to-image translation is challenging, since most of existing face datasets do not contain sufficient images for each age point.
% Simply modeling face aging with CycleGAN~\cite{cyclegan} is not recommended because we should build a large number of deep neural networks as many as the number of target ages.
To alleviate this problem, existing face aging studies~\cite{antipov2017face, palsson2018generative, liu2019attribute, he2019s2gan,caae,IPCGAN} group several continuous ages into discrete age groups (\eg, 20s, 30s, and so on).
% Also, to save the required memory resources, IPCGAN~\cite{IPCGAN} and S2GAN~\cite{he2019s2gan} adopted multi-domain image translation methods~\cite{stargan, he2019attgan} to train with a pair of generator and discriminator~\cite{gan}. 
However, using discrete age groups leaves obvious limitations behind.
Because the model learns only one representative aging factor for each age group, translated results are deterministic in each target age group and thus are discrete, \eg, only one result can be obtained in the target age group of 30s as shown in the Figure~\ref{fig:intro_figure}(a).
Moreover, due to the absence of any clue on the exact age of the translated image, continuous face aging cannot be achieved.
One would try directly applying interpolation between encoded features of a generator, but the interpolated samples are not likely to correspond to the target age because the learned age domains are discrete.

To overcome the limitations, we introduce a novel method to translate a facial image to the continuous target age while preserving the personal identity, named continuous face aging generative adversarial networks (CFA-GAN). 
% In face aging, the feature of personal identity is likely to be entangled with age-related features. 
To learn continuous aging factors, we propose to extract age-invariant personal features by disentangling the features into the identity basis feature and the age basis feature.
To make the decomposed features contain the appropriate information, we train an auxiliary age regressor and an identity classifier by the joint learning strategy.
Moreover, to preserve the original identities of input images, we design a loss maximizing the similarity between the identity basis features of real and translated image.
Consequently, our CFA-GAN is able to generate realistic and smooth images given continuous target ages, as shown in Figure~\ref{fig:intro_figure}(b).

% For face aging with continuous ages, since the target age domain is grouped rather than separated, the continuous face aging is usually conducted by interpolating the features between different target domains. However, since the learned age domains are grouped, interpolated feature is not likely to be the exact representation of target age domain.  We tackle this interpolation method, since the target age is associated with features of the target age group, which means that interpolation should be conducted in intra-domain of the target age group, not between different age groups.

\begin{figure}[tb]
\begin{minipage}[b]{1.0\linewidth}
  \centering
  \centerline{\includegraphics[width=1\textwidth]{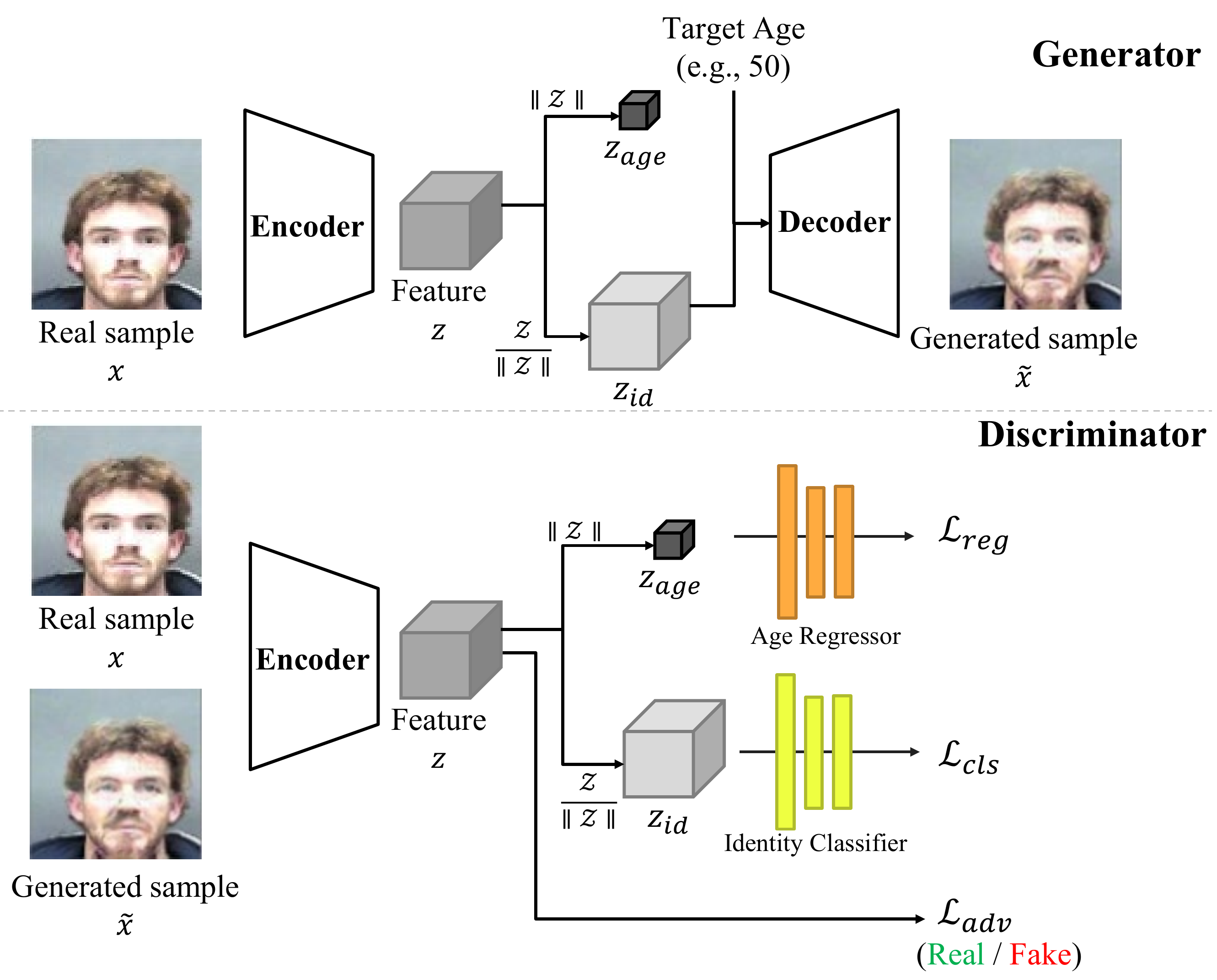}}
  %\vspace{2.0cm}
\end{minipage}
\caption{Overall architecture of CFA-GAN. It consists of a pair of a generator and a discriminator. The encoders in the discriminator and the generator have the same architecture but do not share the parameters. An image feature $z$ is disentangled to the identity basis feature $z_{id}$ and the age basis feature $z_{age}$. Losses of the generator are omitted for brevity.}
\label{fig:architecture}

\end{figure}

% In this paper, we propose to disentangle the features by separating features into the personal basis feature and the age basis feature. With help of the auxiliary feature age classifier, we adversarially train personal basis encoder to encode age-invariant personal basis from the image. We also propose the novel loss function and learning strategy to disentangle age basis feature space, allowing continuous face aging and rejuvenation of the given image. With both qualitative and quantitative evaluations on benchmark face datasets, we demonstrate the continuous aging ability of our model, showing effectiveness of our methods.
% \kb{In this paper, we propose to disentangle the features into the personal basis feature and the age basis feature. With the help of the age regressor, we adversarially train encoder to encode age-invariant personal basis feature maps from an given image. We also propose the novel objective function and the learning strategy to disentangle the age basis feature space, allowing continuous face aging of given image. With both qualitative and quantitative evaluations on benchmark datasets~(MORPH~\cite{morph}), we demonstrate the effectiveness of our methods for continuous aging.}

The main contributions of this paper are as follows.
\begin{itemize}
    \item We present the novel framework that is capable of generating face images of continuous target ages. Note that this is the first attempt to handle continuous target ages.
    \item We propose to disentangle the identity and the age basis features for continuous aging.
    Moreover, a novel loss is designed to preserve the identity of an input image.
    \item We demonstrate the efficacy of our method on continuous aging with the experiments on MORPH~\cite{morph}.
    % \item We demonstrate the effectiveness of proposed method via experiments on the MORPH dataset~\cite{morph}.
\end{itemize}

\section{Proposed Method}
\label{sec:proposed method}

% \subsection{Overall Architecture}

In this section, we introduce the proposed model, Continuous Face Aging Generative Adversarial Network (CFA-GAN). 
As described in Figure~\ref{fig:architecture}, CFA-GAN consists of a generator and a discriminator. The generator is designed based on the U-Net~\cite{unet} architecture with skip-connections.
Each of the generator and the discriminator has an age regressor and an identity classifier.
It is worth noting that their weights are not shared by the generator and the discriminator.

\textbf{Input details.}
The training dataset contains $N$ images $\left\{x^{(i)}\right\}^N_{i=1}$ and the corresponding labels $\left\{y^{(i)}, a^{(i)}\right\}^N_{i=1}$, where $y^{(i)}$ is an identity number and $a^{(i)}$ is an original age.

%------------------------------------------------------------------------
\subsection{Face Aging by Disentangling Age and Identity}

We first extract the feature of an input image using the encoder of the generator $G$. Formally, $z^{(i)} = Enc_{G}(x^{(i)})$, where $x^{(i)}$ and $z^{(i)}$ represent the $i$-th input image and the corresponding personal feature, respectively.
Then, since the personal feature is likely to be entangled with the age-related feature, we propose to decompose the personal feature so that the age-related feature and the age-invariant personal feature are orthogonal. This is formalized as:
\begin{equation}
    \label{equ:decompose}
        z^{(i)} = z_{age}^{(i)} \cdot z_{id}^{(i)},
\end{equation}
where $z_{age}^{(i)} = \left \| z^{(i)} \right \|_{2}$, $z_{id}^{(i)} = \{\frac{z_1}{\left \| z \right \|},\frac{z_2}{\left \| z \right \|},\cdots,\frac{z_c}{\left \| z \right \|} \}$, with $\left \| z_{id}^{(i)} \right \|_{2} = 1 $, $C$ denotes the number of channels and $\left \| \cdot \right \|_{2}$ is the $l2$-norm operator.
The identity basis feature is the age-invariant feature representing the personal identity, while the age basis feature is the age-related features of person.

To ensure that each decomposed feature contains intended information, we take the advantage of multi-task learning. Specifically, we predict the age of an input image with $z_{age}$, and classify the identity number of an input image with $z_{id}$. The loss functions of multi-task learning are as follow.
\begin{equation}
    \label{equ:age_regression}
        \mathcal{L}_{reg} = \frac{1}{N}\sum_{i=1}^{N}(a^{(i)} - f_{age}(z_{age}^{(i)}))^{2}
\end{equation}
\begin{equation}
    \label{equ:id_classify}
        \mathcal{L}_{cls} = - \frac{1}{N}\sum_{i=1}^{N}y^{(i)}\log(f_{id}(z_{id}^{(i)})),
\end{equation}
where $a^{(i)}$ and $y^{(i)}$ denote the ground-truth age and the identity number of the $i$-th input image respectively. The age regressor $f_{age}$ and the identity classifier $f_{id}$ are composed of several fully-connected layers.

Our goal is to translate the aging factor of an input image according to the target age while preserving the personal identity. Hence, after decomposing the feature of an input image, we feed the identity basis feature $z_{id}^{(i)}$ and the target age as inputs to the decoder of the generator. The synthesized images can be obtained by:
\begin{equation}
    \label{equ:img_translation}
        \tilde{x} = Dec( z_{id}, a_{trg} ),
\end{equation}
where $a_{trg}$ denotes the target age.

%------------------------------------------------------------------------
\subsection{Generating Images with Fidelity}

To generate aged images with fidelity, we adopt the adversarial training, following the training process of Generative Adversarial Networks (GAN)~\cite{gan, wgangp, hong2021arrow}. The discriminator is trained not only to discriminate the generated images from the real images, but also to predict their ages and identities. We disentangle features as in Eq.~\ref{equ:decompose}, and adopt multi-task losses as in Eq.~\ref{equ:age_regression} and Eq.~\ref{equ:id_classify}. The multi-task losses are calculated with decomposed features of $Enc_{D}(x)$, where $Enc_{D}$ denotes the encoder of the discriminator.
For quality and stability, we adopt Wasserstein GAN with gradient penalty~\cite{wgangp} as our adversarial loss, which is formulated as:
\begin{equation}
    \label{equ:wganloss}
        \begin{split}
                \mathcal{L}_{adv} & = \underset{\tilde{x} \sim P_g}{\mathbb{E}}[D(\tilde{x})]
                -\underset{x \sim P_r}{\mathbb{E}}[D(x)] \\
                & + {\gamma}\underset{\hat{x} \sim P_{\hat{x}}}{\mathbb{E}}[(\lVert\nabla D(\hat{x})\rVert_{2}-1)^{2}],
        \end{split}
\end{equation}
where $D$ denotes the discriminator. With the adversarial training, the generator learns to produce realistic results.

The translated image $\tilde{x}$ should also be estimated same as the target age while preserving the personal identity. Hence, we adopt an age error loss of fake images to optimize the generator. The age error loss is the mean squared error (MSE) between the estimated age and the target one, defined as:
\begin{equation}
    \label{equ:age_error}
        \mathcal{L}_{age} = \frac{1}{N}\sum_{i=1}^{N}(a_{trg} - f_{age}(\tilde{z}^{(i)}_{age}))^{2},
\end{equation}
where $a_{trg}$ indicates the randomly sampled target age.

To preserve the original personal identity, we propose to minimize the verification loss by maximizing the cosine similarity between the identity basis feature of the original and translated images as follows.
\begin{equation}
    \label{equ:id_preservation}
        \mathcal{L}_{id} = 1 - \frac{z_{id} \cdot \tilde{z}_{id}}{\left \| z_{id} \right \| \left \| \tilde{z}_{id} \right \|}.
\end{equation}
where $\tilde{z}_{id}$ denotes the identity basis feature of the generated image $\tilde{x}$.
The identity preservation loss is minimized when the angle between two feature vectors are 0.

Lastly, to train without ground-truth, we adopt the reconstruction loss and the cycle consistency loss~\cite{cyclegan} as follows.
\begin{equation}
    \label{equ:reconstruction}
        \mathcal{L}_{recon}=\frac{1}{N}\sum_{i=1}^{N}(x^{(i)} - G(x^{(i)},a^{(i)}))^{2}.
\end{equation}
\begin{equation}
    \label{equ:cycle_consistency}
        \mathcal{L}_{cycle}=\frac{1}{N}\sum_{i=1}^{N}(x^{(i)} - G(G(x^{(i)},a_{trg}),a^{(i)}) )^{2},
\end{equation}

The overall loss functions of the discriminator and the generator are as follows.
\begin{equation}
    \label{equ:overall_D}
        \mathcal{L}_{D} = \mathcal{L}_{adv} + \lambda^{D}_{reg}\mathcal{L}_{reg} + \lambda^{D}_{cls}\mathcal{L}_{cls}.
\end{equation}
\begin{equation}
    \label{equ:overall_G}
    \begin{split}
        \mathcal{L}_{G} = - \mathcal{L}_{adv} + \lambda^{G}_{reg}\mathcal{L}_{reg} + \lambda^{G}_{cls}\mathcal{L}_{cls} + \lambda_{age}\mathcal{L}_{age} \\ + \lambda_{id}\mathcal{L}_{id} + \lambda_{recon}\mathcal{L}_{recon} + \lambda_{cycle}\mathcal{L}_{cycle},
    \end{split}
\end{equation}
where $\left\{\lambda_{*}\right\}$ are hyper-parameters for weighing loss functions.

\begin{figure}[tb]

\begin{minipage}[b]{1.0\linewidth}
  \centering
  \centerline{\includegraphics[width=0.9\textwidth]{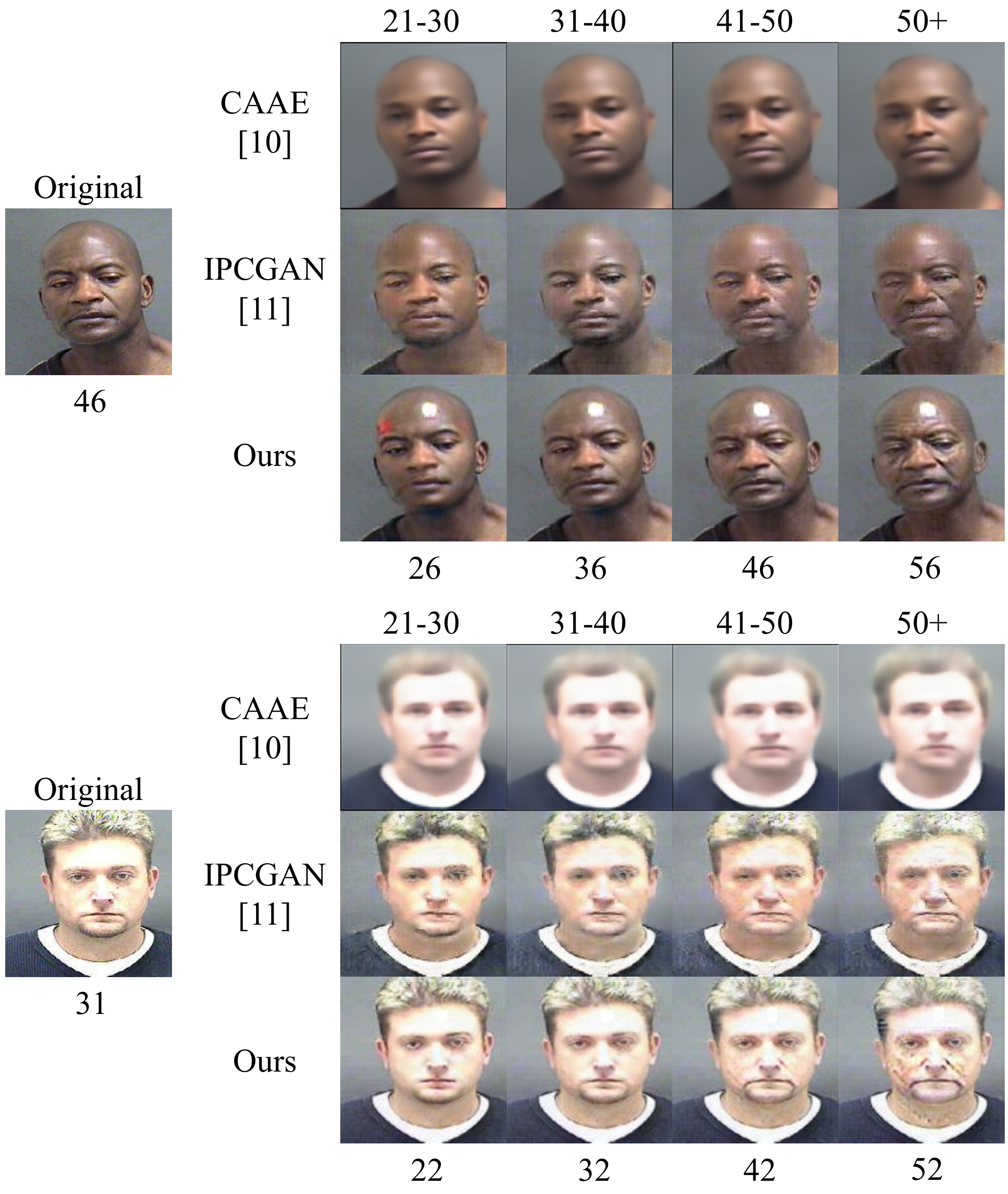}}
%  \vspace{2.0cm}
\end{minipage}
\caption{Qualitative results on face aging/rejuvenation. The ground-truth ages are denoted below the original images. We report the translation results of the original age and the target ages with intervals of 10 below the result images. Our method generates results better representing target ages than other methods.}
\label{fig:quality_comparison}%
\end{figure}
%--------------------------------------------------------------------------------------------------------------
\section{Experiments}
\label{sec:experiments}
In this section, we provide the implementation details of our CFA-GAN and report evaluation results on the MORPH~\cite{morph} dataset. MORPH contains 55,000 face images of 13,617 identities from 16 to 77 years old.
Following the prior works~\cite{caae,IPCGAN,AcGAN}, we first extract facial regions of 200 $\times$ 200 pixels using MTCNN~\cite{MTCNN},
% pre-process the dataset by cropping 200 x 200 pixels of facial region detected by MTCNN~\cite{MTCNN},
and then resize them to 128 $\times$ 128 resolution.
% Since our method aims to translate an input image with continuous target age, we do not divide age labels into age groups unlike prior works.
We split the dataset into training and test set in a ratio of 90:10 respectively. To demonstrate the effectiveness of our CFA-GAN, comparative experiments are conducted with the state-of-the-art face aging methods.

\subsection{Implementation Details}
% Following Pix2Pix~\cite{pix2pix}, we adopt UNet~\cite{unet} architecture with skip connections as our generator. Specifically, our generator is an encoder-decoder architecture.
Before the feature decomposition (Eq.~\ref{equ:decompose}), we apply the global average pooling.
The target age condition is normalized to be in the range of $[-1,1]$ based on the minimum and the maximum ages of the dataset.
% Age regressors and identity classifiers are shallow Multi-layer Perceptron~(MLP) models. 
Weighting factors of loss functions are set as follows.
$\left\{\lambda^{D}_{reg}, \lambda^{G}_{reg}\right\}=0.001$, $\left\{\lambda^{D}_{cls}, \lambda^{G}_{cls}\right\}=0.1$,
$\lambda_{age}=0.02$, $\lambda_{id}=1$, $\lambda_{recon}=10$, and $\lambda_{cycle}=10$.
The discriminator and the generator are alternately trained by the Adam~\cite{Adam} optimizer with the learning rate of $10^{-4}$. The batch size is set to 16.
All experiments are conducted on a single 1080 Ti GPU.

\begin{table}[]
% \small
\footnotesize
\resizebox{1.0\columnwidth}{!}{%
\begin{tabular*}{\columnwidth}{l@{\extracolsep{\fill}}cccc}
\hline
\multicolumn{5}{c}{Estimated Age Distribution}                        \\ \hline
Age group~$\rightarrow$      & 21-30       & 31-40       & 41-50       & 50+         \\ \hline
Generic        & 25.12       & 35.43        & 44.72         & 54.88         \\
CAAE~\cite{caae}    & 24.31 & 31.02 & 39.03 & 47.84 \\
IPCGAN~\cite{IPCGAN} & 22.38 & 27.53 & 36.41 & 46.42 \\
AcGAN~\cite{AcGAN}         & 25.92 & 36.49 & 40.59 & 47.88 \\
Ours           & 26.88 & 36.96 & 48.85 & 59.28 \\ \hline
\multicolumn{5}{c}{Estimated Age Error}                                \\ \hline
Ours           & 4.58  & 6.22  & 7.35  & 7.02  \\ \hline
\end{tabular*}%
}
\caption{Evaluation on the regressed ages of translated images. ``Generic'' denotes the ground-truth age distribution. We report the average estimated age distributions and the mean squared errors between the target and the estimated ages.}
\label{tab:eval_age}
\end{table}

\subsection{Qualitative Results}

We qualitatively compare our method with the state-of-the-art face aging methods~\cite{caae, IPCGAN}. As shown in Table~\ref{fig:quality_comparison}, our CFA-GAN successfully translates input images to the designated ages while preserving original identities. Since previous methods are trained with discrete age group labels, face aging to the original age group often fails to reconstruct the input image. On the contrary, with the ability of continuous face aging, our method well reconstructs original images without undesired transformation. 

In terms of identity preservation, both CAAE~\cite{caae} and IPCGAN~\cite{IPCGAN} fail to preserve the original identities of input images. For instance, some facial attributes tend to be changed or lost. On the other hand, with the help of the identity preservation loss ($\mathcal{L}_{id}$), most of the facial attributes are preserved and only necessary parts are translated.

\subsection{Quantitative Results}
Face aging aims to translate the faces of input images to target ages while preserving personal identities. Therefore, one can evaluate face aging models from two perspectives: (1) how much the age of the translated image matches the target age and (2) how well the identity of the original image is preserved. Following the previous studies~\cite{IPCGAN, AcGAN}, we evaluate our CFA-GAN by employing Face++ API~\cite{faceplusplus} that offers age regression and face verification.
% As in previous methods~\cite{IPCGAN, AcGAN}, we evaluate our method quantitatively with the help of Face++ API~\cite{faceplusplus}. The first evaluation metric is the aging accuracy, which is the difference between the estimated age of a translated image and the target age label. Since existing studies are trained and tested with age group labels, we compare by reporting the results of 5 randomly sampled target ages in each age group. We also report the mean and the variance of age differences for fair comparison. Another evaluation metric is the face verification rate, which evaluates how well the personal identity is preserved in the translated image.

\begin{table}[t]
\footnotesize
%\resizebox{\columnwidth}{!}{%
% Please add the following required packages to your document preamble:
% \usepackage{graphicx}
\begin{tabular*}{\linewidth}{c@{\extracolsep{\fill}}cccc}
\hline
\multicolumn{5}{c}{Verification Confidence (\%)}                            \\ \hline
\multicolumn{1}{c|}{Age group} & 21-30   & 31-40   & 41-50   & 50+     \\ \hline
\multicolumn{1}{c|}{21-30}     & 94.12   & 93.71   & 90.97   & 87.49   \\
\multicolumn{1}{c|}{31-40}     & 93.36   & 93.90   & 92.61   & 89.40   \\
\multicolumn{1}{c|}{41-50}     & 92.15   & 93.58   & 92.98   & 90.52   \\
\multicolumn{1}{c|}{50+}       & 89.82   & 92.39   & 92.56   & 91.01   \\ \hline
\multicolumn{1}{c|}{Model}     & \multicolumn{4}{c}{Verification Rate (\%)} \\ \hline
\multicolumn{1}{c|}{CAAE~\cite{caae}}      & 99.38   & 97.82   & 92.72   & 80.56   \\
\multicolumn{1}{c|}{IPCGAN~\cite{IPCGAN}}    & 100     & 100     & 100     & 100     \\
\multicolumn{1}{c|}{AcGAN~\cite{AcGAN}}     & 100     & 100     & 100     & 100     \\
\multicolumn{1}{c|}{Ours}      & 100     & 99.93     & 99.68     & 98.91 \\ 
\hline
\end{tabular*}%
%}
\caption{Evaluation on identity preservation. The upper part presents the average confidence scores of the translated images from the original age (row) to the target age (column). The lower part shows verification rates, where a pair is considered true positive if its confidence surpasses the threshold.}
\label{tab:eval_verification}
\end{table}

%----------------------------------------
Firstly, to measure how close the age of the generated image is to the target age, we report the estimated age distributions and estimated age errors using Face++ in Table \ref{tab:eval_age}.
As shown in the upper part, all the results of our CFA-GAN fall into the corresponding target age groups while those of all the other methods do not in some groups, \eg, 41-50 and 50+.
In addition, since our work is capable of dealing with continuous target ages, we report the age error (in the lower part) which is the mean of absolute differences between the actual target ages and the estimated ages of translated images.
% Estimated age errors shows that most of CFA-GAN produced samples are within target age groups.
We observe that the age errors are relatively high when the target ages are larger than 30, probably due to the data imbalance.
This problem could be alleviated by improving the age regressors in our framework, which is one of our future directions.

%----------------------------------------

% \begin{figure}[tb]

% \begin{minipage}[b]{1.0\linewidth}
%   \centering
%   \centerline{\includegraphics[width=8.5cm]{figure/qualitative_ablation.pdf}}
% %  \vspace{2.0cm}
% \end{minipage}
% \caption{Qualitative ablation study.}
% \label{fig:quality_ablation}
% %
% \end{figure}

%----------------------------------------
Secondly, we evaluate our method in the aspect of identity preservation.
% We give a pair of an original image and its translated image to Face++, then the API produces the confidence score for whether the pair contains the same person.
The upper part of Table~\ref{tab:eval_verification} presents the average confidence scores of Face++ given the translated images from the original age (row) to the target age (column).
It can be noticed that the results of our method show high verification confidences when compared with their original images. Meanwhile, the lower part of Table 2 shows the verification rates, where a pair is considered true positive if its confidence surpasses the pre-defined threshold and false positive otherwise. Following the convention, we set the threshold to 73.795 \% with a small error rate of 0.001 \%. Our method achieves high verification rates for all age groups.

\section{Conclusion}
\label{sec:conclusion}
Since previous face aging studies were limited in discrete age group labels, the ages of their result images were in black box. In this paper, we proposed CFA-GAN for continuous face aging, where the age-related features are isolated from the age-invariant features. Moreover, we proposed a novel loss function to preserve original personal identity. As a result, the translated images by our method correspond well to the target ages (not groups). Through the experiments, we validated the superiority of our method against the existing work.
% \vfill\pagebreak

% References should be produced using the bibtex program from suitable
% BiBTeX files (here: strings, refs, manuals). The IEEEbib.bst bibliography
% style file from IEEE produces unsorted bibliography list.
% -------------------------------------------------------------------------
{
\bibliographystyle{IEEEbib}
\bibliography{refers}
}
\end{document}